\documentclass[conference]{IEEEtran}
\IEEEoverridecommandlockouts
\usepackage{cite}
\usepackage{amsmath,amssymb,amsfonts}
\usepackage{algorithmic}
\usepackage{booktabs}
\usepackage{graphicx}
\usepackage{textcomp}
\usepackage{siunitx}
\usepackage{tablefootnote}
\usepackage[symbol]{footmisc}

\usepackage{xcolor}
\usepackage{hyperref}
\def\BibTeX{{\rm B\kern-.05em{\sc i\kern-.025em b}\kern-.08em
    T\kern-.1667em\lower.7ex\hbox{E}\kern-.125emX}}
\begin{document}


\title{\rule{\textwidth}{2pt}\\Imposing Connectome-Derived Topology on an Echo State Network\\\vspace{-.5cm} \rule{\textwidth}{2pt}} 




\author{\IEEEauthorblockN{\textbf{Jacob Morra}\textsuperscript{1}}
\IEEEauthorblockA{
Department of Computer Science\\
The Brain and Mind Institute\\
The University of Western Ontario\\
London, ON Canada \\
\texttt{\href{mailto:jmorra6@uwo.ca}{jmorra6@uwo.ca}}}

\and

\IEEEauthorblockN{\textbf{Mark Daley}\textsuperscript{2}}
\IEEEauthorblockA{
Department of Computer Science\\
The Brain and Mind Institute\\
The University of Western Ontario\\
London, ON Canada \\
\texttt{\href{mailto:mdaley2@uwo.ca}{mdaley2@uwo.ca}}}
}

\IEEEaftertitletext{\vspace{-.8cm}}

\maketitle

\begin{abstract}
Can connectome-derived constraints inform computation? In this paper we investigate the contribution of a fruit fly connectome's \textit{topology} on the performance of an \textit{Echo State Network} (ESN) --- a subset of \textit{Reservoir Computing} which is state of the art in chaotic time series prediction. Specifically, we replace the \textit{reservoir} layer of a classical ESN --- normally a fixed, random graph represented as a 2-d matrix --- with a particular (female) fruit fly connectome-derived connectivity matrix. We refer to this \textit{experimental} class of models (with connectome-derived reservoirs) as ``Fruit Fly ESNs'' (FFESNs). We train and validate the FFESN on a chaotic time series prediction task; here we consider four sets of trials with different training input sizes (small, large) and train-validate splits (two variants). We compare the validation performance (Mean-Squared Error) of all of the best FFESN models to a class of \textit{control} model ESNs (simply referred to as ``\textit{ESNs}''). Overall, for all four sets of trials we find that the FFESN either significantly outperforms (\textit{and} has lower variance than) the ESN; or simply has lower variance than the ESN.  \end{abstract}

\vspace{-.1cm}
\section{Introduction}

Reservoir Computers (RC) were first introduced as easily-trainable Recurrent Neural Networks; however, they have since grown to occupy a niche in chaotic time-series prediction. For this task type, RCs yield best-in-class performance whilst requiring less training time and memory \cite{b1}. In the space of Machine Learning models, RCs can be thought of as a class of predictors which resemble biological brains \cite{b2}; notably, this resemblance \textit{requires some squinting}.
\\

Much of the recent work on \textit{Brain-Inspired Artificial Intelligence} has been ``top-down'', applying some conceptual understanding of biological intelligence to a machine learning approach \cite{b3,b4,b5}. Here we instead wish to consider a ``mesoscale'' approach from early principles; specifically in that we will derive a particular aspect of our RC architecture from a \textit{connectome} (a ``map'' of brain connections). We call our approach one from ``early principles'' in that we opt to select the \textit{simplest} biological brain which retains a set of desirable capabilities (for a predictor) --- in particular, efficiency and discriminatory ability. Intuitively, this leads us to consider the olfactory system neural network of the fruit fly.
\\

In considering olfaction alone, the fruit fly (Drosophila Melanogaster) is both \textit{effective} and \textit{efficient}. For the former, it \textit{has to} be effective at odor discrimination for survival and mating: among other behaviours, it can selectively target odors from fruit metabolism \cite{b6}; it can also classify pheromone odors of potential mates to determine their age \cite{b7}. For the latter, it also needs to be efficient at odor discrimination (again, for its own survival): as an example, it identifies new odors in one-shot, ``tagging'' new scents with an accuracy similar to those it has observed hundreds of times (\cite{b8,b9}). 
\\

How can we insert a connectome into a Reservoir Computer? The key bridging factor here is the RC's \textit{reservoir}: a random, fixed graph of nodes which transforms inputs into a non-linear higher dimensional space. Since the reservoir can be represented as a 2-d matrix, here we can generate and insert a \textit{connectivity matrix} (see Methodology) from the connectome in its place. For clarity, since we are only concerned with imposing connectome topology, we will only consider the non-spiking class of RCs: ``Echo State Networks'' (ESNs). The basic ESN architecture is illustrated below (Figure 1). To train an ESN, discrete time series input is fed into the network and allowed to ``echo'' around the reservoir; ultimately, a simple linear regression (with an L2 regularization penalty) minimizes the error between the target and predicted outputs. For a more formal description of the ESN, please see ``Methodology''.

\begin{figure}[htbp]
\centerline{\includegraphics[width=80mm]{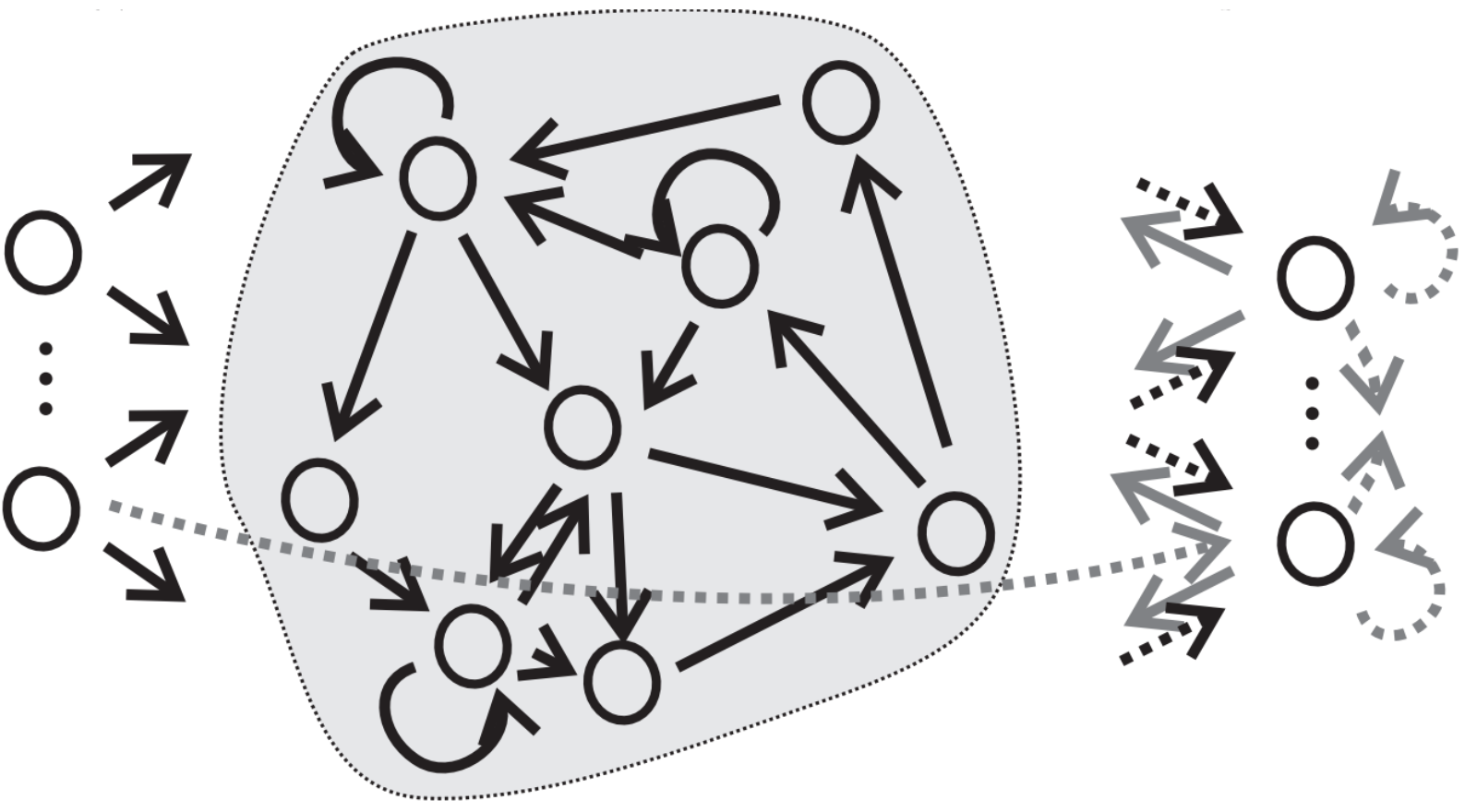}}
\caption{The Echo State Network architecture (adapted from \cite{b10}, Fig. 1).}
\label{fig1}
\end{figure}

In our previous work, we created three connectivity matrices: one for each of the major ROIs (regions of interest) of the olfactory system --- the antenna lobe, the mushroom body calyx, and the lateral horn. We then implanted these matrices into a Multi-layer Perceptron (MLP) with three hidden layers (one for each ROI); here the ROI neuron counts were used to specify the number of nodes in each corresponding hidden layer, and the connectivity matrices were used as the weight vectors between layers \cite{b11}. This approach was limiting for two reasons: First, the discrete structure of the hidden layers prevented within-layer connections (thus many weights were not included); Second, because hidden layer weights would change during training, the connectome-derived weights were not preserved. The ESN addresses both of these issues: First, the entire connectivity matrix (i.e. as opposed to three separate connectivity matrices) is preserved within the reservoir; Second, the reservoir weights are \textit{fixed} (only the readout layer of the network is trainable).
\\

Our contribution in this work is a set of performance comparisons between a \textit{control model ESN} and an \textit{experimental model ESN} --- one which has had its reservoir topology constrained by that of the fruit fly connectome; we call this latter model the \textbf{``Fruit Fly ESN'' (FFESN)}. Ultimately, we predict that the FFESN will achieve superior performance to the control model on one or more variants of a particular time-series prediction task (see Methodology). 
\\

In the next section (``Methodology'') we will further describe the process of generating a connectivity matrix from the fruit fly connectome; we will also formally describe the ESN, the time series prediction task (and variants) considered, our particular model classes, and our process for selecting optimal hyperparameters. In the ``Results and Discussion'' section, we will identify the best selected models (from hyperparameter optimization) and present and discuss the performance of these models. Finally, in the ``Conclusion'' section we will summarize the results, present the project limitations, and highlight directions for future work.
\\

\section{Methodology}
\subsection{Generating the Connectivity Matrix}

Here we describe the process of creating our connectivity matrix, which is used to impose a topological constraint on our ESN reservoir by forcing a particular, connectome-derived structure, as opposed to a randomly-generated one. For the connectome we consider Janelia's hemibrain \cite{b12}. The hemibrain is the largest and most complete fruit fly connectome reconstruction to date. It is an Electron-Microscopy-derived 3-d volume (from a particular female subject) which comprises over 25,000 neurons, each grouped by a specific cell type and in their respective brain regions of interest. To retrieve the hemibrain data we first access the API through a generated key; next, we query the hemibrain database to select all neurons and connections within the olfactory system (the antenna lobe, mushroom body calyx, and lateral horn); we then store all neuron body IDs, all connections (i.e. from each neuron body ID to all other neuron body IDs it is connected to), and all ``weights'' between neurons. Here, \textit{weight} is the number of synapse-synapse connections between two neurons (as originally described in \cite{b12}). From this point, we select only those neurons which reside in the right lateral horn: this is a simplification which we have opted for in order to reduce training times of our control and experimental models. 
\\

With all connectivity information (neuron body IDs plus the weights between all other neuron body IDs) for the right lateral horn in hand, we generate the \textbf{connectivity matrix} (see Figure 2). This $n$x$n$ matrix --- where $n$ is the number of right lateral horn neurons --- identifies all connections \textit{from} a particular body ID (indicated by the row index, as in Figure 2) to a particular body ID (indicated by the column index). As previously mentioned, the cell values in this matrix correspond to the number of synapse-synapse connections between neurons (the weights). Under an assumption that each neuron is ``connected to itself'', the diagonal of the weight matrix is filled with ones. In Figure 2 we illustrate a small subsection of the (normalized) connectivity matrix.
\\

\subsection{The Echo State Network}
Here let us formally describe the network dynamics of the ESN. Given a discrete-time input signal $\mathbf{u}(n) \in \mathbb{R}^{N_u}$ and target signal $\mathbf{y}(n) \in \mathbb{R}^{N_y}$, the goal is to learn $\hat{\mathbf{y}}(n) \in \mathbb{R}^{N_y}$ such that an error measure $E\bigg(\hat{\mathbf{y}}(n), \mathbf{y}(n)\bigg)$ is minimized \cite{b2}. The recurrent reservoir dynamics (as previously specified in \cite{b2}) can be described by the following:
\begin{equation}
    \tilde{\mathbf{x}}(n) = \mathit{tanh}\bigg(\mathbf{W}^{in}[1;\mathbf{u}(n)]+\mathbf{Wx}(n-1)\bigg)
\end{equation}
\begin{equation}
    \mathbf{x}(n) = (1-\alpha)\mathbf{x}(n-1)+\alpha \tilde{\mathbf{x}}(n)
\end{equation}

Here $[\cdot;\cdot]$ is a column vector; $\mathbf{W}^{in}$ is a weight vector between the input and reservoir layers ($\mathbf{W}^{in}\in \mathbb{R}^{N_x \times (1+N_u)}$);

\begin{figure}[htbp]
\centerline{\includegraphics[width=90mm]{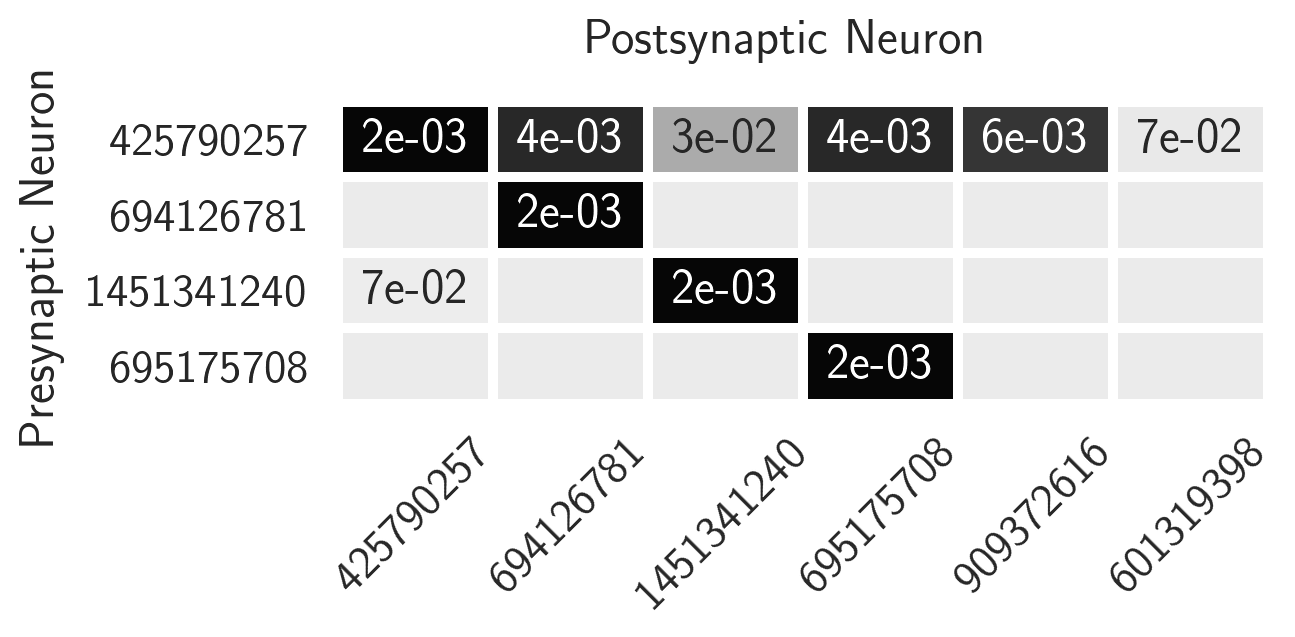}}
\caption{A \textbf{subset} of the connectivity matrix used for this work which illustrates the number of synapse-synapse connections (normalized)  between neurons in the fruit fly connectome (the hemibrain). Blank entries are zero-valued.}
\label{fig1}
\end{figure}

\noindent $\mathbf{W}$ is the reservoir weight vector ($\mathbf{W} \in \mathbb{R}^{N_x \times (N_x)}$); $x(n)$ contains the reservoir activations ($x(n) \in \mathbb{R}^{N_x}$); $\tilde{\mathbf{x}}(n)$ contains the updated activations; $\alpha$ is the \textit{leaking rate}, ($\alpha \in (0,1]$). 
\\

\noindent The ESN is trained only at the readout (output) layer. The readout layer is defined as:
\begin{equation}
    \mathbf{y}(n)=\mathbf{W}^{out}[1;\mathbf{u}(n);\mathbf{x}(n)]
\end{equation}
\noindent We can rewrite this in matrix notation:
\begin{equation}
    \mathbf{Y} = \mathbf{W}^{out}\mathbf{X}
\end{equation}

\noindent A standard approach is to solve this system with L2-penalized regression (Ridge Regression): 
\begin{equation}
    \mathbf{W}^{out}=\mathbf{Y}\mathbf{X}^{T}\bigg(\mathbf{X}\mathbf{X}^{T}+\lambda\mathbf{I}\bigg)^{-1}
\end{equation} 

\noindent Here $\lambda$ is the \textit{regularization coefficient} \cite{b2}.
\\

\subsection{Time Series Prediction}
For a time series prediction task we consider the Mackey-Glass dataset (Figure 3) \cite{b13}. This chaotic, 1-d time series (describing physiological control systems) is a conventional benchmarking dataset for comparison of ESNs \cite{b14}. We generate this data from the following ODE \cite{b15}:
\\
\vspace{-.25cm}

\begin{equation}
\frac{dx}{dt}=\beta x(t)+\frac{\gamma x(t-\tau)}{1+x(t-\tau)^{10}}
\end{equation}
\\
\vspace{-.25cm}

Here, $x(t)$ is the value of the Mackey-Glass time series at time $t$, and $[\gamma, \beta, \tau]$ are some real-valued constants. The objective of the time series prediction task for a series of size $T$, given $x(t)$ for $t$ $\in$ $[0,1,...,m-1]$, is the correct prediction of $x(t)$ for $t$ $\in$ $[m+r,m+r+1,...,T]$. Here $r$ controls how far into the future our model predicts; $m \in \mathbb{Z^{+}}$ and $r \in \mathbb{Z^{+}}$.
\\

Consider a subset of the Mackey-Glass dataset: specifically $x(t)$ for $t \in [0,T]$. We define the \textit{training input} for this dataset (on a time series prediction task) as the subset $[0,A], A << T$; here we consider $A=300$ and $A=900$. For these two training input sizes we will also select from two train-validate split variants: we will refer to these as \textbf{``Variant A''} and \textbf{``Variant B''}. The \textit{Variant A} (as in \cite{b16}) train-validate split follows a train-validate-train-validate ordering whereby the training input time series is followed by the validation input time series, and then the training output time series is followed by the validation output time series. Here the inputs and outputs for each set must be the same sizes; moreover, we select the majority of the data for training (input and output). The \textit{Variant B} train-validate split follows a train-train-validate-validate ordering, where training output follows training input and validation output follows validation input. Figure 4 illustrates these split variants. A distinction between the two variants lies in their perceived difficulty; specifically, for a particular training input size $A$, we note that Variant A requires prediction of values which are far into the future as compared to Variant B --- we therefore would expect poorer performance from both models for the former variant.
\\

\subsection{Two Classes of Models}
We consider two classes of models: the Fruit Fly Echo State Network (FFESN) and the Echo State Network (ESN). Both are identical with the exception of the topology of their \textit{reservoirs}. The \textbf{ESN} (as previously defined in subsection B) serves as a \textbf{control model}: it has a fixed, random reservoir (a 2-d matrix) with its values drawn from a uniform distribution over $[0,1)$. As in \cite{b16}, a mask of 0's is applied to impose a sparsity of 20\%; the locations of 0 elements in the mask are determined by comparing the designated reservoir density ($0.2$) with a 2-d array filled with values sampled from a uniform distribution over $[0,1)$. To ensure that our ESN reservoir is \textit{fixed} through all iterations, we enforce a random seed; for added measure, we also store, re-assign, and display this array after each iteration (i.e. for all predictions). The \textbf{FFESN} (our \textbf{experimental model}) has a fixed reservoir derived from a connectivity matrix, as described in subsection A. To implement all models we use the \textit{easyesn} package \cite{b16}.
\\

\subsection{Hyperparameter Selection}
Given the FFESN and ESN classes of models, for each of the training input sizes and train-validate split variants (described in subsection C of this section) we wish to identify the best-performing models (of each class) for each task. We therefore optimize the hyperparameters $[\lambda, \alpha]$ for each task-specific model class using a \textit{grid search} with $\lambda \in [0,1]$, $\alpha \in (0,1]$. The leaking rate, $\alpha$, determines the contribution of the previous and current reservoir activations on the current reservoir activation (see Equation 2). The regularization coefficient, $\lambda$, determines the magnitude of the Ridge Regression penalty (see Equation 5). Notably, we have opted out of selecting for other hyperparameters such as the \textit{spectral radius} as these would alter our experimental and control reservoir topologies.
\\

\begin{figure}[htbp]
\centerline{\includegraphics[width=90mm]{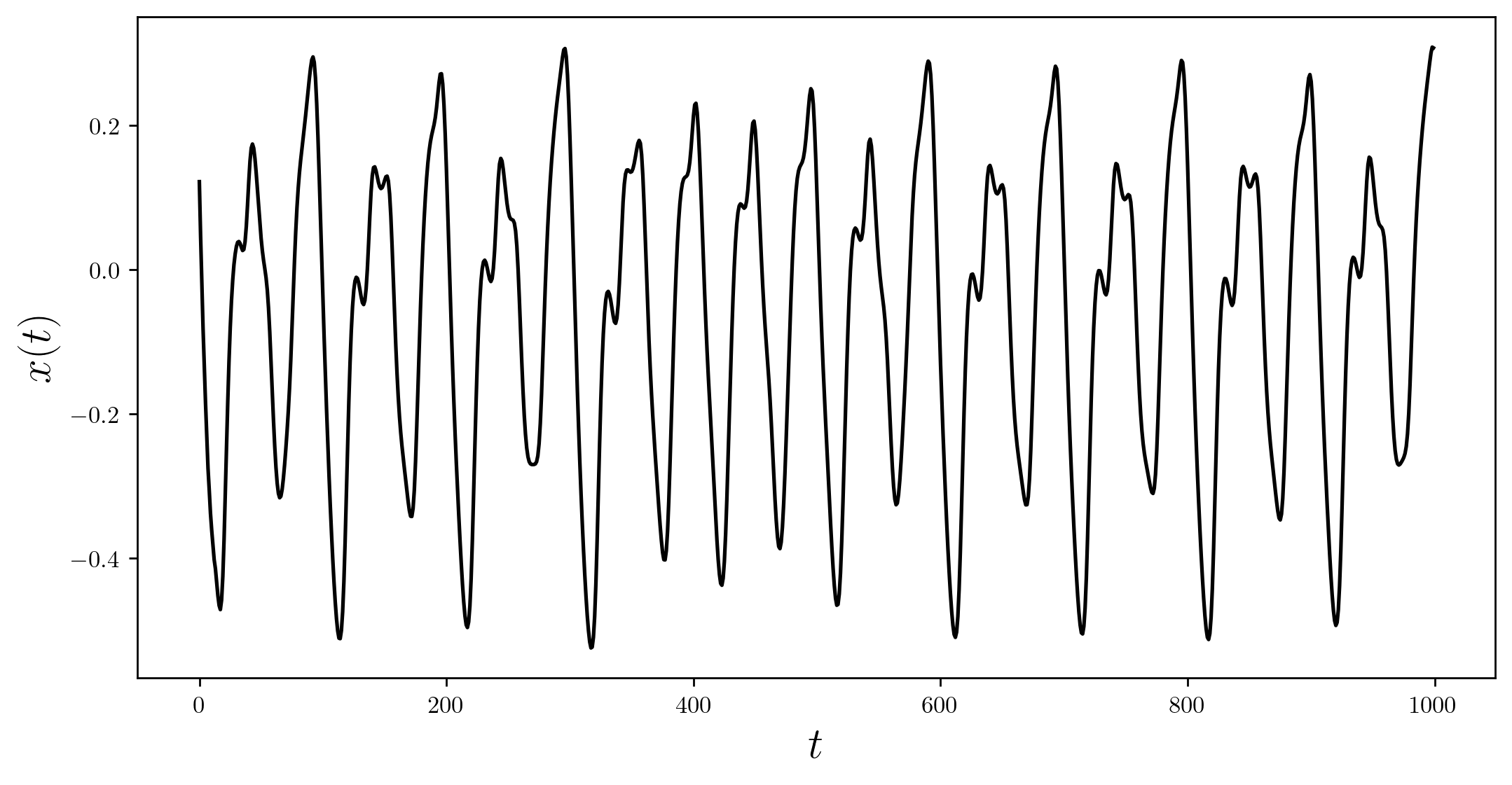}}
\caption{The ``Mackey-Glass'' chaotic time series.}
\label{fig1}
\end{figure}

For each training input size (900, 300) and train-validate split variant (A, B), we train and validate a population of models with varying hyperparameters. From this population of models we then select the best performers and choose their specific combination of hyperparameters for each \textit{class} of model (FFESN, ESN). For a given split variant and training input size, we then repeatedly train and validate our chosen models 50 times in order to generate 50 time series predictions. Finally, for a performance metric we consider the Mean-Squared Error (MSE) between each best-model prediction time series and the ground truth time series; we report a 95\% Confidence Interval (CI) around each MSE (see Table I). The Mean-Squared Error is defined as the following:

\begin{equation}
\label{eqn:MSE}
\text{MSE} = \frac{1}{n}\sum_{i=1}^{n}(y(t_{i})-\hat{y}(t_i))^2,
\end{equation}
where $n$ is the number of time steps; $y(t_i)$ is the labelled output at time $t_i$, and $\hat{y}(t_i)$ is the predicted output at time $t_i$.
\\

\section{Results and Discussion} 
\subsection{Time Series Prediction (Variant A)}
For Variant A we consider training input sizes $A=900$ and 300. For $A=900$ we select the first 2000 discrete time series points from the Mackey-Glass dataset. We assign training input and output sets as $x(t)$ for $t$ $\in [0:900]$ and $[1000:1900]$, respectively; validation input and output sets are $x(t)$ for $t$ $\in [900:1000]$ and $[1900:2000]$. We retain this train-validate sequence for $A=300$ and again use 200 time series points in the validation set (the rest is used for training). After performing hyperparameter optimization, for $A=900$ we select $[\alpha, \lambda] = [0.1, 0.0024]$ for the FFESN and $[\alpha, \lambda] = [0.2, 4.53$E-$05] $ for the ESN; for $A=300$ we select $[\alpha, \lambda] = [0.1, 0.0024]$ for the FFESN and $[\alpha, \lambda] = [0.2, 0.0024] $ for the ESN.
\\

Figures 5-8 illustrate our best-model predictions and prediction errors on the validation set for 50 train-validate iterations; here we consider training input sizes of 900 and 300 (Variant A). In Figure 5 (top subfigure) we observe (for $A=900$) that the FFESN better captures the two peaks in the original data. In the bottom subfigure of Figure 5 we observe that the ESN CI is much wider than the FFESN CI. Intuitively, these results are congruent with the error-based plots in Figures 6 and 7: In Figure 6, it is apparent that for $A=900$ the FFESN MSE is much lower than the ESN MSE; the distribution of errors is also more narrow. In Figure 7, for $A=300$ again we see a tighter distribution of errors with the FFESN, but also report similar MSEs. Figure 8 elucidates our findings from the directly preceding figures: in summary, for 50 train-validate iterations the FFESN \textit{significantly} outperforms the ESN for $A=900$; importantly, the FFESN also yields a fraction of the ESN's variance (in MSE) for both training input sizes ($A=900$ and $A=300$).  
\\

\begin{figure}[htbp]
\centerline{\includegraphics[width=95mm]{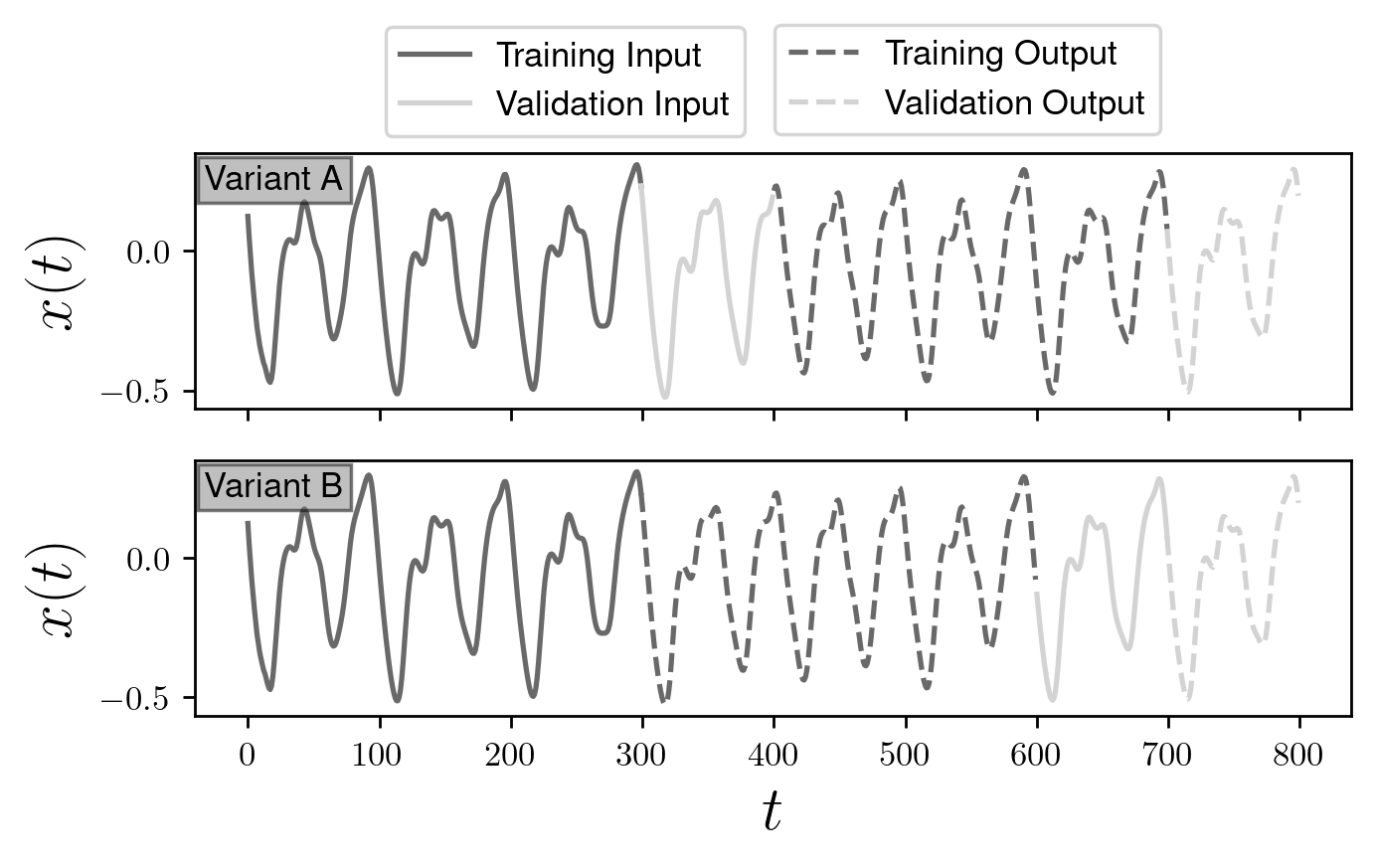}}
\vspace{-0.5cm}
\caption{The two variants of train-validate split considered in this work (in this case, for a training input size of $A=300$).}
\label{fig1}
\end{figure}

\begin{figure}[htbp]
\centerline{\includegraphics[width=90mm]{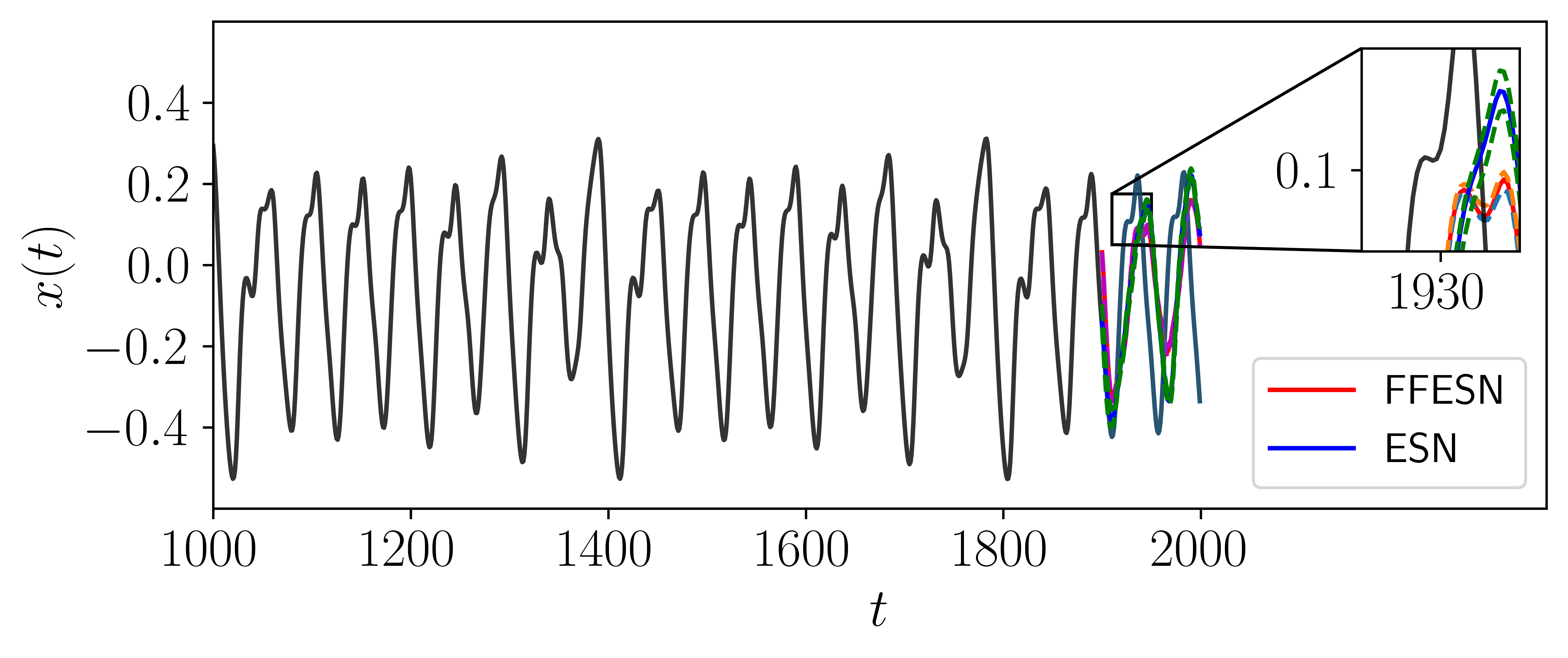}}
\hfill
{\includegraphics[width=90mm]{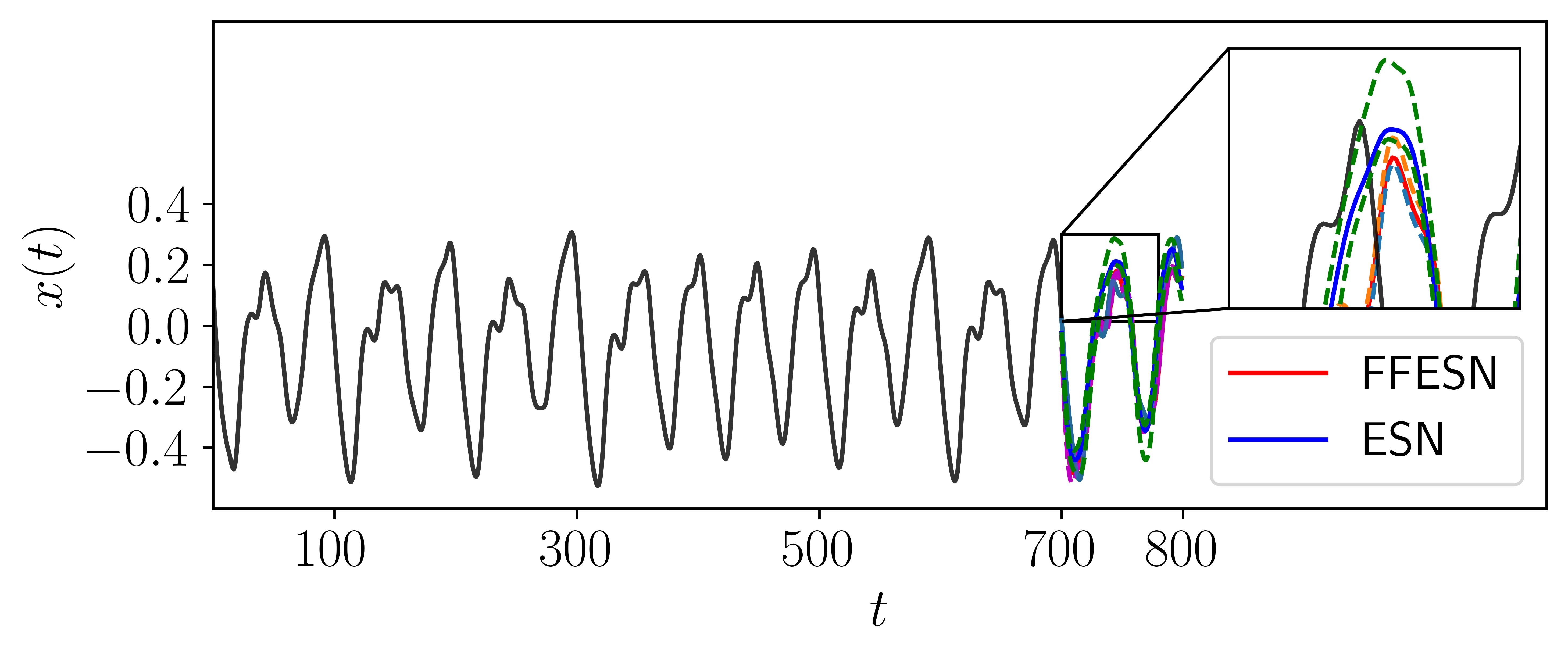}}
\vspace{-0.8cm}
\caption{Time series predictions (Variant A) for the last 100 steps of a portion of the Mackey-Glass dataset (for $A=900$ and 300, respectively from top to bottom). The green and orange dotted lines around the prediction curves (blue and red lines) denote their 95\% Confidence Interval margins.}
\end{figure}

\begin{figure}[htbp]
\centerline{\includegraphics[width=80mm]{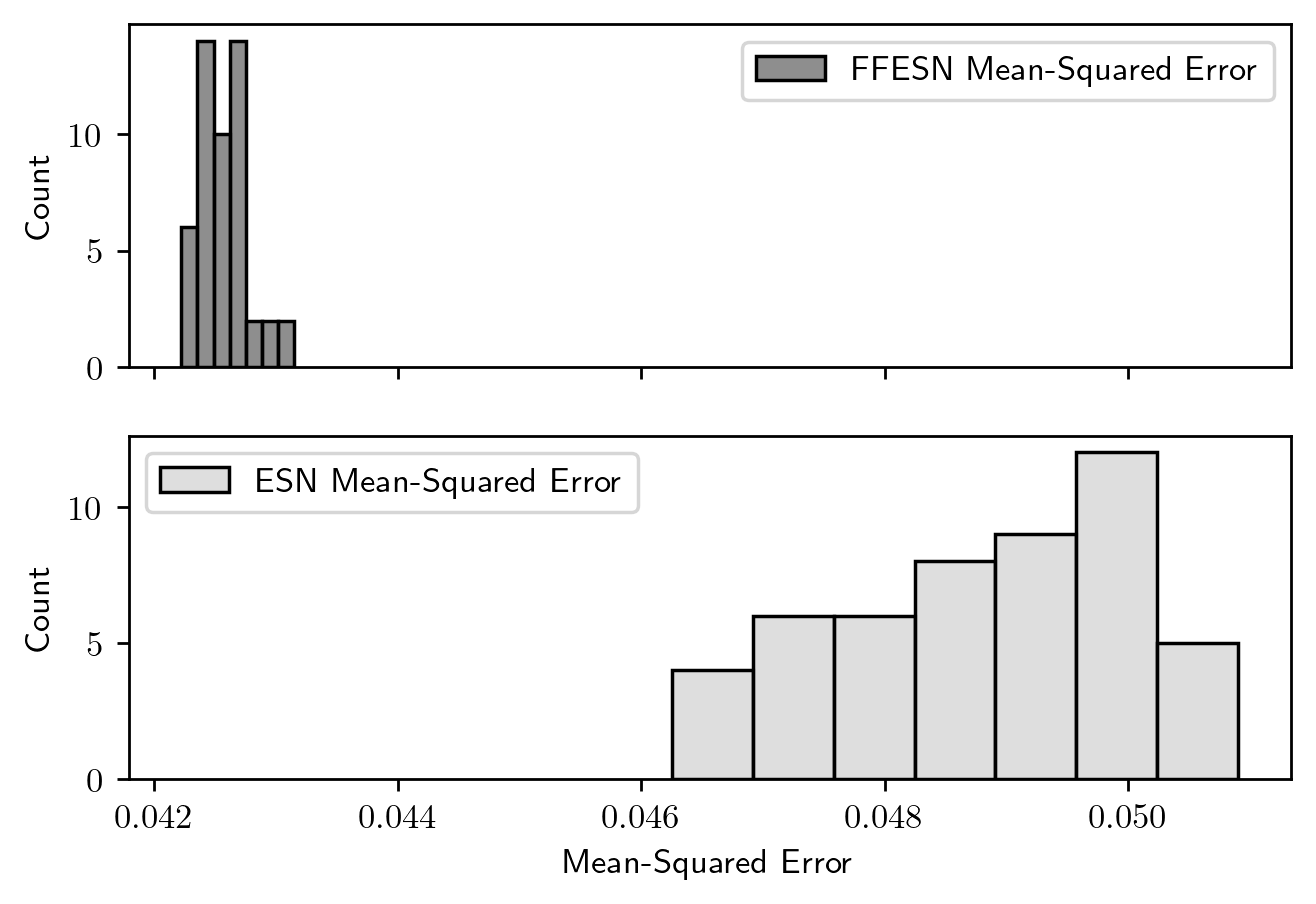}}
\caption{A histogram illustrating the Mean-Squared Error for 50 iterations of best-model (FFESN and ESN) predictions on the validation set (Variant A). Here we consider a training input size of 900 time steps.}
\label{fig1}
\end{figure}

\begin{table*}[]
\caption{Summary of Comparisons between Experimental (FFESN) and Control (ESN) Model Classes}
\begin{tabular}{@{}lcccccccc@{}}
\toprule & \multicolumn{4}{c}{\textbf{$\mathbf{A=900}$ (training input size)}} & \multicolumn{4}{c}{\textbf{$\mathbf{A=300}$ (training input size)}}                                              \\ 
\midrule
\multicolumn{1}{l|}{} & \multicolumn{2}{c|}{\textit{Variant A}} & \multicolumn{2}{c|}{\textit{Variant B}} & \multicolumn{2}{c|}{Variant A} & \multicolumn{2}{c}{Variant B}                        \\
\multicolumn{1}{l|}{} & FFESN & \multicolumn{1}{c|}{ESN} & FFESN & \multicolumn{1}{c|}{ESN} & FFESN & \multicolumn{1}{c|}{ESN} & FFESN & ESN            
\\
\multicolumn{1}{l|}{\textit{Best $\lambda$}} & 0.0024 & \multicolumn{1}{c|}{4.53E-05} & 0.00127 & \multicolumn{1}{c|}{0.0357} & 0.0024 & \multicolumn{1}{c|}{0.0024} & 0.189 & 1           
\\
\multicolumn{1}{l|}{\textit{Best $\alpha$}} & 0.1 & \multicolumn{1}{c|}{0.2} & 0.2 & \multicolumn{1}{c|}{0.6} & 0.1 & \multicolumn{1}{c|}{0.2} & 0.1 & 0.2                 \\
\multicolumn{1}{l|}{\textit{MSE 95\% CI}} & \textbf{{[}0.042, 0.043{]}} & \multicolumn{1}{c|}{{[}0.046, 0.051{]}} & {[}0.016, 0.017{]} & \multicolumn{1}{c|}{{[}0.013,0.025{]}} & {[}0.0048,0.0057{]} & \multicolumn{1}{c|}{{[}0.0022, 0.0067{]}} & \textbf{{[}0.0041, 0.0042{]}} & {[}0.0049, 0.0062{]} 
\\
\multicolumn{1}{l|}{\textit{MSE Variance}} & \textbf{3.76E-08} & \multicolumn{1}{c|}{1.46E-06} & \textbf{4.89E-08}  & \multicolumn{1}{c|}{9.03E-06} & \textbf{5.21E-08}   & \multicolumn{1}{c|}{1.24E-06} & \textbf{1.36E-09} & 1.03E-07             
\\ 
\bottomrule
\end{tabular}
\centering
\scriptsize
\tablefootnote[2]{} Bold values indicate the (significantly) best-performing or lowest-variance model for particular train-validate split variants. 
\end{table*}

\subsection{Time Series Prediction (Variant B)}
For the Variant B train-validate split again we consider large ($A=900$) and small ($A=300$) training input sizes. As in Variant A, for $A=900$ we select the first 2000 Mackey-Glass data points. In contrast to Variant A, here the first 900 discrete time steps are for training input and the next 900 steps are for training output; following this, the next 100 steps and remaining 100 steps are reserved for the validation input and output, respectively. This train-train-validate-validate split is maintained for $A=300$, with the only difference being that there are 800 time series points in total (instead of 2000). After performing hyperparameter optimization, for a training input size of 900 we select $[\alpha, \lambda] = [0.2, 0.00127] $ for the FFESN and $[\alpha, \lambda] = [0.6, 0.0357] $ for the ESN; for 300 training inputs we select $[\alpha, \lambda] = [0.1, 0.189] $ for the FFESN and $[\alpha, \lambda] = [0.2, 1] $ for the ESN. 
\\

Figure 9 provides an illustration of the prediction performance of the best FFESN and ESN models for 50 train-validate iterations on the validation set for \textit{Variant B}. In the figure we observe that for 900 training inputs, whilst there is no significant difference in prediction performance, the variance of the FFESN is a \textit{fraction} of that of the ESN. For a training input size of 300, we observe that the FFESN outperforms the ESN \textit{significantly} and also has a fraction of the MSE variance. Overall, this complements the results for the Variant A train-validate split, suggesting that the FFESN either outperforms \textit{and} has lower variance than the ESN, or simply has lower variance than the ESN. 
\\

Table I summarizes the results from Figures 5-9. It also includes the optimal hyperparameters for each model class on a particular train-validate split (i.e. A or B) with a certain training input size (900 or 300). Importantly, from this table it is evident that \textit{for all of the train-validate split variants and training input sizes} we observe a \textbf{hundred-fold decrease} in MSE variance for the FFESN class of models (as compared to the ESN class of models). We also observe that the FFESN significantly outperforms the ESN (outside of two standard deviations of the mean) in the $A=900$, Variant A trials and $A=300$, Variant B trials. 
\\

From our results, the observed reduction in variance from the FFESN is surprising. It is particularly of importance as 

\begin{figure}[htbp]
\centerline{\includegraphics[width=80mm]{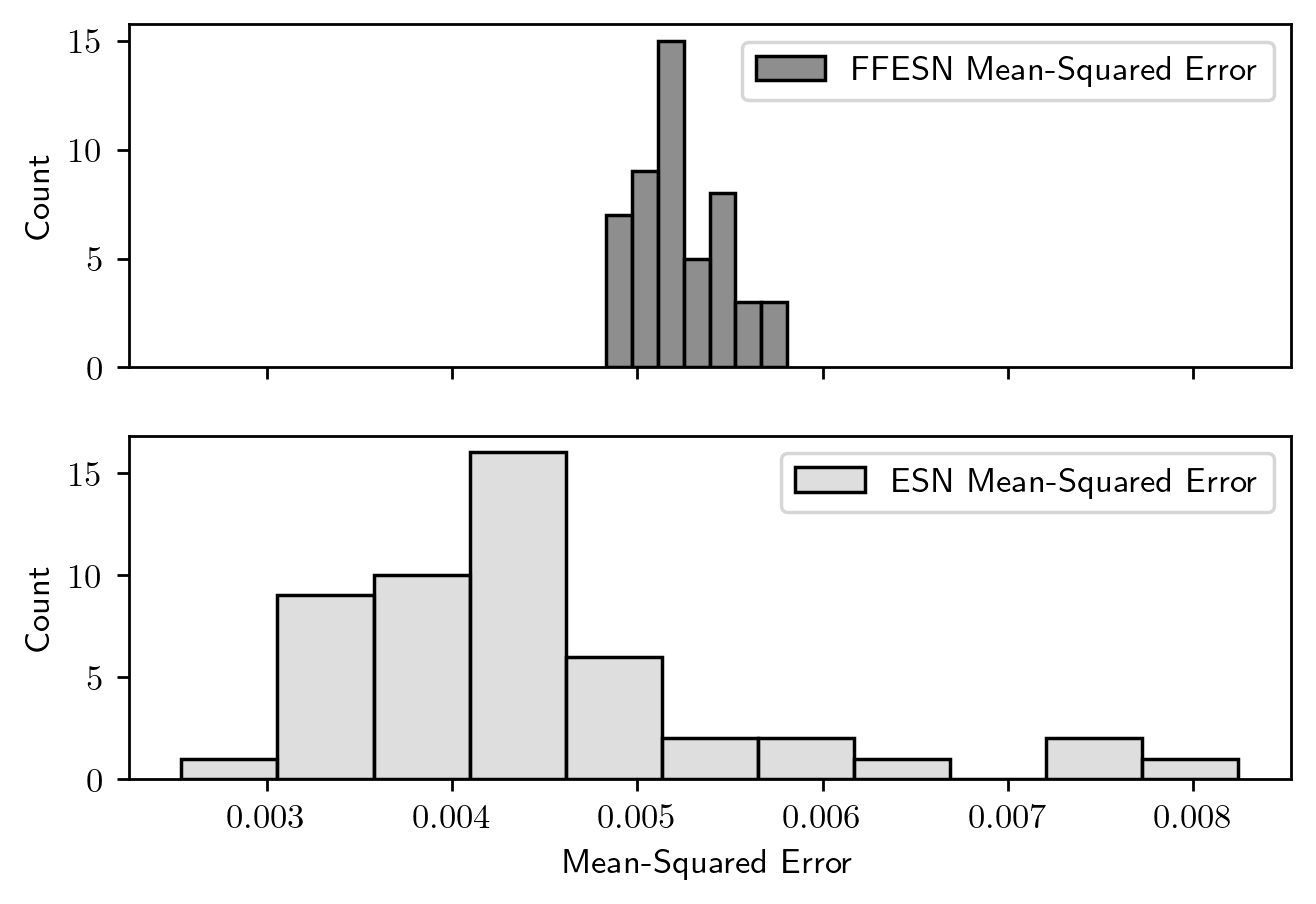}}
\caption{A histogram showing the MSE for 50 iterations of best-model (FFESN and ESN) predictions on the validation set (Variant A). For this plot we have used a training input size of $A=300$ time steps.}
\label{fig1}
\end{figure}

\begin{figure}[htbp]
\centerline{\includegraphics[width=80mm]{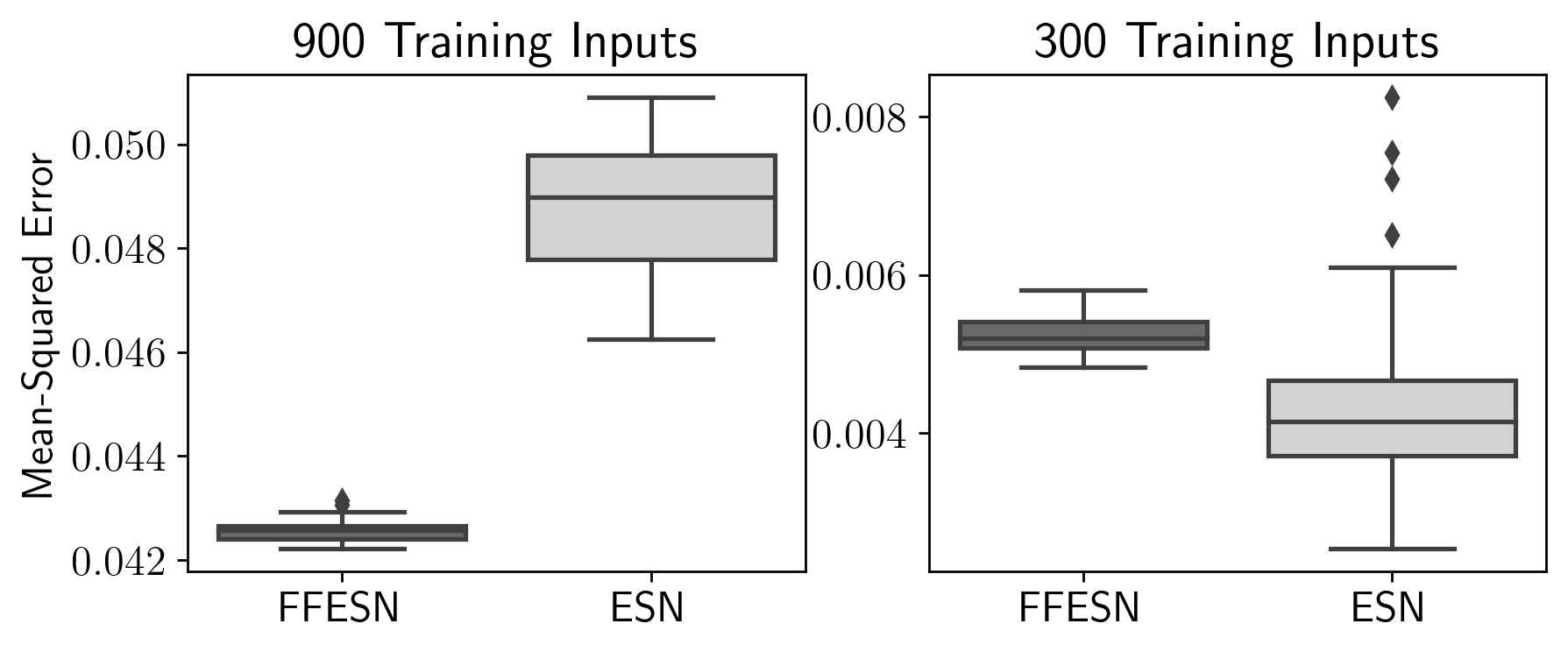}}
\caption{For Variant A, these boxplots show the Mean-Squared Error for 50 trials of predictions on the validation set for the FFESN and ESN (training input sizes are $A=900$ and $A=300$).}
\label{fig1}
\end{figure}

\begin{figure}[htbp]
\centerline{\includegraphics[width=80mm]{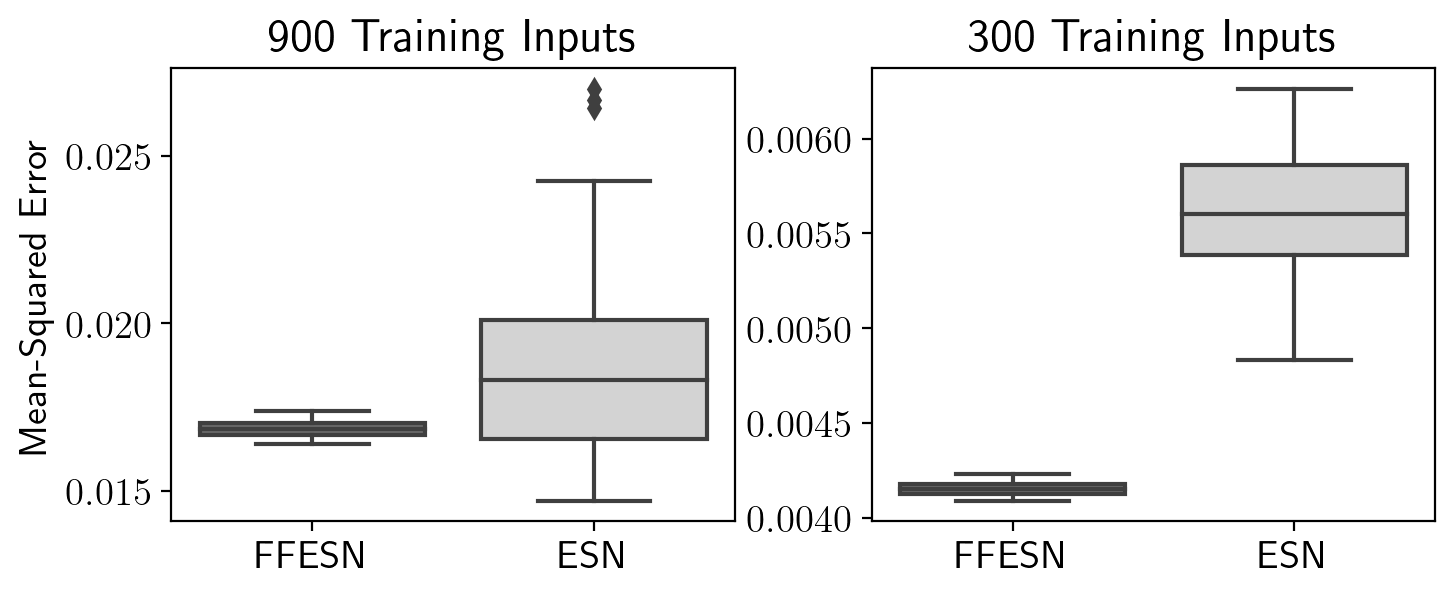}}
\caption{For Variant B, these boxplots show the MSE for 50 trials of validation predictions for the FFESN and ESN with training input sizes of 900 and 300.}
\label{fig1}
\end{figure}

\noindent high variance is a known challenge for ESNs (\cite{b17, b18}). To consider a potential explanation for this outcome, we intuit that the fruit fly brain retains a topological structure which (likely as a result of selective evolutionary pressures) favours consistent output responses (i.e. low variance); this idea is not unlike current theories regarding brain modularity as a result of evolutionary pressure \cite{b19}; however, at this stage it is purely speculative and remains an open question.

\section{Conclusion}
In summary, we have provided experimental evidence supporting a significant (100-fold) reduction in performance variance in conjunction with similar or improved performance metrics after imposing brain-derived topological restrictions on an Echo State Network. Specifically, after hyperparameter optimization of the experimental (FFESN) and control (ESN) classes of models on two train-validate split variants --- each with two training input sizes considered --- we have observed that the FFESN either significantly outperforms the ESN with greatly reduced variance; or yields comparable performance (not significantly different) to the ESN while maintaining a significant reduction in variance. With our results we have validated the initial proposition that the FFESN would outperform the ESN in a time series prediction task --- that is, for a particular dataset, training input size, and split variant. \\

To extend our previous discussion above, it is apparent that some particular aspect of the topological structure of the fruit fly olfactory brain is capable of contributing to a reduction in ESN variance (and a performance increase); however, we do not yet know which specific feature is responsible. As a next step, we therefore propose an investigation into \textit{three structural components} of the fruit fly neural network and their contributions to ESN performance (and variance): \textit{small-worldness} (i.e. the clustering coefficient) \cite{b20}, \textit{sparsity}, and network \textit{weights}. Here we would create three experimental models (one for each aspect). As another direction for future work, a study on the impact of connectome topology on ESN \textit{training efficiency} would be worth considering; primarily because the fruit fly is an efficient learner (\cite{b8,b9}); but also in light of recent work in the Reservoir Computing space on ``Next Generation Reservoir Computers'', where \cite{b1} reports similar performance to conventional Reservoir Computers but with reduced computational requirements (shorter training time, lower data requirements).
\\

Speaking to one of the inherent limitations of this work, it should be noted here that whilst all of the right lateral horn neurons are present in the hemibrain, future iterations of this work --- wherein we would hope to use the complete olfactory neural system --- would be missing a portion of neurons which make up the network topology; for example, only 83\% of right antenna lobe neurons are mapped within the hemibrain \cite{b12}. To mitigate this we will strive to only consider ROIs with some tolerated level of trace completion.
\\

Overall, with this work we have presented a particular set of results which can serve as motivation for looking towards biological brains to yield performance improvements in a given class of machine learning models; moreover, from our results we wish to facilitate further lines of comparison between biologically-motivated, topologically-constrained machine learning models and their conventional counterparts.

\end{document}